\documentclass{article}
\usepackage{graphicx}
\usepackage{longtable}
\usepackage{amsmath}
\usepackage{multirow}
\usepackage{csquotes}
\usepackage{tikz}
\usepackage{float}
\usepackage{hyperref}
\usepackage{url}

\usetikzlibrary{arrows.meta,positioning}
\usetikzlibrary{calc}

\usepackage[
backend=biber,
style=apa,
sorting=nyt
]{biblatex}
\usepackage{geometry}
\title{Benchmark of stylistic variation in LLM-generated texts}
\author{Jiří Milička, Anna Marklová, Václav Cvrček\footnote{E-mails: jiri@milicka.cz; anna.marklova@ff.cuni.cz; vaclav.cvrcek@ff.cuni.cz. All authors are affiliated with the Department of Linguistics, Faculty of Arts, Charles University, Prague. This document is an initial draft of the manuscript prepared before submission for review.}}
\date{}

\begin{document}

\maketitle

\section*{Abstract}

This study investigates the register variation in texts written by humans and comparable texts produced by large language models (LLMs). Biber's multidimensional analysis (MDA) (\cite{Biber1988a}) is applied to a sample of human-written texts and AI-created texts generated to be their counterparts to find the dimensions of variation in which LLMs differ most significantly and most systematically from humans. As textual material, a new LLM-generated corpus AI-Brown is used, which is comparable to BE21 (a Brown family corpus representing contemporary British English). Since all languages except English are underrepresented in the training data of frontier LLMs, similar analysis is replicated on Czech using AI-Koditex corpus and Czech multidimensional model (\cite{Cvrcek.etal2018}). Examined were 16 frontier models in various settings and prompts, with emphasis placed on the difference between base models and instruction-tuned models. Based on this, a benchmark is created through which models can be compared with each other and ranked in interpretable dimensions.

\bigskip
\textbf{Keywords}: English; Czech; Large Language models; factor analysis; multidimensional analysis; stylistic variation

\section{Introduction}
Companies developing frontier Large Language Models (LLMs), while proclaiming their aim toward general intelligence \cite{altman2023planning, amodei2024machines}, are in practice creating systems designed to score highly on benchmarks focused on mostly technical tasks (particularly programming, problem solving, and instruction following). The training methodology and the selection of training data are subordinated to this goal, often training directly on public benchmarks \cite{balloccu2024leak, ni2025training}. These decisions are both marketing-driven (results on hard science benchmarks present well) and pragmatic, since technically proficient professionals are a source of revenue. Moreover, these companies are primarily software developers, so it is natural that they create tools that would help themselves (\cite{garg2025google}). Technically proficient AI is also necessary to initiate the cycle of recursive self-improvement.

This way we cease to view the language models as a language models in the original sense of the term, as used for decades in Computational Linguistics and Natural Language Processing. Instead of a next-word predictor, it becomes a problem-solving agent. It appears, however, that the original base model, which is trained on very diverse texts, becomes flattened through this additional training and loses the ability to simulate stylistically and cognitively diverse personas that are not this problem-solving \emph{helpful assistants}. This observation is, however, only subjective and would need to be tested, which is the main goal of this article. Specifically, we focus on stylistics-related research questions:

\begin{enumerate}
    \item How well can current LLMs produce stylistically diverse texts from various genres and text types?
    \item Are texts created using current LLMs stylistically shifted  consistently across different models? I.e., is there some AI-language stylistic attractor?
    \item How do instruction-tuned models differ from base models?
    \item What is the difference between texts generated using a simple system prompt and texts generated using long \emph{helpful assistant} system prompt?
    \item Are stylistic features dependent on the sampling temperature?    
    \item Is the stylistic shift smaller in English than in a language underrepresented in the training data?
\end{enumerate}

To explore these topics, we utilize two corpora of LLM-generated texts: AI Brown (English) and AI Koditex (Czech) (\cite{Milicka2025AIBrown, Milicka2025AIKoditex}), which are stylistically very diverse. English language was selected as the primary language on which frontier LLMs are trained. However, English is exceptional in this regard, since LLMs are trained on vastly more English data than data from other languages. For this reason, we also used a Czech corpus, which better represents medium-sized languages. Both corpora were created using similar methodology, with the main goal of being directly comparable to human-written texts: original text chunks from Koditex \cite{Zasinaetal2018} and BE21 (Brown family corpus) \cite{baker2023year} are divided into two parts, where the first serves as a prompt for the LLM to continue, and the second part serves as a reference text to determine whether the LLM continued stylistically consistently.

Stylistic consistency is measured stylometrically using standard Biber's Multi-Dimensional Analysis (MDA) methodology (\cite{Biber1988a, nini2019multi}), which has been adapted for Czech by Cvrček (\cite{Cvrcek.etal2018}). The approach involves measuring a large number of stylometric features, which are subsequently reduced using factor analysis. The advantage of this approach is that the resulting dimensions are already interpreted, for example the second dimension in English has been interpreted as \emph{narrativity}, so if LLM-generated texts have shifted in this dimension, we can say that the text is more narrative than the original.

This method leads to the creation of a stylistic benchmark through which models can be compared on several well-interpretable dimensions. In our opinion, it would be beneficial for LLM creators to adopt this benchmark to become part of the standard toolkit to assess performance of their models. This benchmark is not only good for determining whether a model is suitable for users who collaborate with LLMs to write various genres of fiction, it shows how the model will affect users more broadly, because as it turns out, humans consider LLM-based systems as friends, therapists, even confidants, not only as problem-solving oracles or assistants \cite{xu2025bonding}.

\subsection{Broader context of the study}

The previous studies on the stylometry of LLMs have primarily focused on developing automatic human / AI classifiers, or AI detection systems. Recent studies have demonstrated various approaches to distinguishing AI-generated content, with accuracy rates ranging from 81\% to 98\% using stylometric features (\cite{kumarage2023stylometricdetectionaigeneratedtext, mikros2023ai, mikros2025beyond, Przystalski_2026}). However, our study approaches stylometry in different aims in mind. While AI detection systems examine differences in style in order to exploit them, we look at them to evaluate and interpret the quality of the LLM.

The current consensus appears to be that there is no single distinct AI language, e.g., ChatGPT and Gemini default personas exhibit their own unique writing styles, akin to human idiolects (let us coin the term \emph{aidiolects} for AI idiolects) (\cite{rudnicka_2025, bitton2025detecting, rudnicka2025each, rao2025two}). At the same time, LLMs can simulate many different personas than the default ones (\cite{shanahan2023role,Milicka2024large}) and these personas produce stylistically diverse texts (\cite{malik-etal-2024-empirical}).

Nevertheless, we can identify certain stylistic attractors --- patterns that all or at least most LLMs tend to gravitate toward. This is what actually makes all these human / AI classifiers mentioned in the first paragraph of this section possible. These attractors are driven by similar training data largely derived from Common Crawl (\cite{commoncrawl}) and other easy to get sources, and of course by the architecture itself (\cite{vaswani2017attention}). Furthermore, it is clear that some personas cannot be simulated by LLM-based systems because they are simply missing from their latent space --- either because they are not represented in training data (such as an illiterate person, see \cite{da2024illiterate}) or because they are deliberately suppressed in subsequent post-training (because of political extremism or political views inconsistent with the values of the companies that train these models, see \cite{rozado2024political}). 

How should one characterize these underlying stylistic biases or attractors? Rather than cataloging individual quirks (like overusing certain words or constructions or other stylometric features, like in \cite{reinhart2025llms}), it is useful to adopt a holistic, corpus-linguistic approach. In the tradition of Biber’s Multi-Dimensional Analysis (MDA) \cite{Biber1988a}, we consider that co-occurring linguistic features often reflect an underlying stylistic dimension. If an LLM-generated text contains less past-tense verbs and far less third-person pronouns than a human reference, it’s likely not a coincidence --- those features are both indicators that the model's style is shifted towards less narrativity. These two features (among many others) co-occur commonly in fiction and storytelling genres (which often involve past events and third-person characters) and are jointly absent in, let us say, legal contracts, so they formed a \emph{Narrativity} dimension in Biber’s factor analysis.

Crucially, these dimensions are interpretable: instead of saying a text has 50\% more second person pronouns, 30\% extra present tense verbs, disproportionate number of  ``private'' verbs (\emph{ assume, believe}...), and shorter words on average (which is hard to contextualize), we can say it scores high on the Involved dimension (pointing to an informal, interactive style), for example. Such dimensions have been corroborated by follow-up studies in corpus linguistics and shown to align with our intuitive perceptions of genre differences \cite{biber2019multi}.

\section{Data}
\subsection{Language choice}

When selecting languages and corpora for our experiments, we considered both practical and theoretical factors related to LLMs. English was chosen for several reasons. First, the vast majority of training data for LLMs is in English (e.g. 93 \% in GPT-3, \cite{brown2020languagemodelsfewshotlearners}), and their internal representations are therefore primarily shaped by this language \cite{zhong2024beyond,schut2025multilingual}. Moreover, most engineering decisions during model development were made with English in mind, including the design of tokenization methods such as Byte Pair Encoding (BPE), which are optimized for English \cite{zhong2024beyond}. In this sense, English represents the model’s ``default mode.”

For our experiments, we chose the well-established Brown family corpus BE21 \cite{baker2023year}. While we cannot rule out the possibility that this corpus was included in the models’ training data, this does not pose a problem: the training data is compressed in such a way that models do not memorize original texts verbatim. Our inspection of generated continuations confirmed this --- none of them reproduced the original corpus texts exactly.


\subsubsection{Czech}
As explained earlier, one of our aims was to examine how models perform in a ``smaller'' language, that is, one not centrally represented in their training data. In fact, this applies to virtually all languages other than English. According to \cite{Johnson2022}, as of 2019, the second most represented language in GPT-3’s training data was French (1.8\%), followed by German (1.5\%). All other languages were represented at below 1\%. When studying the language capacities of LLMs, it is therefore crucial to focus on languages other than English, since English occupies a very unique position.

We chose Czech for two practical reasons. First, a pipeline for register analysis developed by Cvrček \cite{Cvrcek.etal2018} allows us to apply Multidimensional Analysis (MDA) to newly generated AI corpora. Second, we had access to the multi-genre Koditex corpus \cite{Zasinaetal2018}, which was used for the original MDA on Czech and therefore provided a well-suited basis for generating comparable AI corpora.

From Koditex, we selected only original Czech texts (excluding translations), omitted poetry, restricted the sample to written texts, and took only one excerpt from each text.

\subsection{Models}

For our study, we selected models that were among the most widely used at their time of release and that represent different approaches to training and ``alignment.”  

A key distinction is between so-called \emph{base} models and \emph{instruction-tuned} models. Base models are raw language models that generate text based solely on training data, without additional optimization for following instructions or for safety. Instruction tuning (often combined with RLHF – reinforcement learning from human feedback) equips models to better follow user prompts, but also alters their linguistic profile.  

The most important base model in our corpus is \textbf{davinci-002}, which was the last available model of this family that had not undergone reinforcement learning. Unlike more recent base models (e.g., Llama or Mistral), for which training on outputs of other language models cannot be excluded, davinci-002 represents a relatively ``clean” pre-alignment example. In this sense, it can be compared to ``pre-nuclear steel,” a scientific term for uncontaminated material produced before atomic testing.  

The \textbf{Llama} (version 3.1, 405 billion parameters, 16 bite quantization) in both base and instruction tuned variant was also included in our sample for English, but did not produce viable texts for Czech.

We also included \textbf{GPT-3.5-turbo}, \textbf{GPT-4}, \textbf{GPT-4-turbo}, \textbf{GPT-4.5}, and \textbf{GPT-4o}, because they were the most widely used OpenAI models at the time of the study conduction and they currently set the standard for large language model applications. A special case is \textbf{GPT-3.5-turbo in completion mode}, which, although not a true base model (it too underwent RLHF), was one of the few models available at the time that supported Czech generation without requiring a system prompt.  

\textbf{Claude 3.5 Haiku} and \textbf{Claude 3.5 Opus}, and \textbf{Claude 4 Sonnet}) were included because they rely on a different alignment method ``constitutional AI”) rather than RLHF, and from the outset attracted considerable attention for its distinctive output characteristics.  

Finally, we added \textbf{Gemini} (1.5 and 2.0, Flash and Pro) and \textbf{DeepSeek v3}, as they represent competitive approaches that were becoming increasingly relevant alternatives to OpenAI and Anthropic models at the time of the study.

\subsection{Sampling temperature}

In the study, we systematically tested two settings of the \emph{temperature} parameter, which controls the degree of randomness in text generation:  

\begin{itemize}
  \item \textbf{0} – a deterministic setting, where the model always produces the same output for the same input.  
  \item \textbf{1} – the original probability distribution, which best reflects the model’s natural variability of outputs.  
\end{itemize}

Additionally, we included the temperature of \textbf{0.5} for GPT-3.5 turbo – to experiment with a compromise value between determinism and variability, allowing for some diversity in output while maintaining relative consistency. 

Some models did not produce viable texts for zero temperature in Czech language (just repeated a single sentence or paragraph indefinitely), so they had to be omitted for this setting.

\subsection{Corpus generation procedure and prompts}
\begin{figure}[ht]
    \centering
\begin{tikzpicture}[
  >=Stealth,
  font=\sffamily,
  box/.style={draw=black, very thick, align=center, inner sep=10pt},
  smallbox/.style={draw=black, very thick, align=center, inner sep=8pt},
  every node/.style={inner sep=6pt}
]

\node[box] (half) at (0,0) {First part of\\the original text};
\node[box] (orig) at (3.5,0) {Second part of\\the original text};

\node[smallbox] (prompt) at (0,-3.5) {``Generate\\ continuation\\ of this text''};

\node[box] (ai) at (3.5,-3.5) {New second part\\ (AI text)};

\draw[black, dash pattern=on 5pt off 4pt, line width=0.88pt] 
  ($(half.north east)!0.5!(orig.north west)$) -- 
  ($(half.south east)!0.5!(orig.south west)$);

\draw[-{Stealth[length=3.5mm,width=2.6mm]}, line width=1.2pt] (half.south) -- (prompt.north);
\draw[-{Stealth[length=3.5mm,width=2.6mm]}, line width=1.2pt] (prompt.east) -- (ai.west);

\node (bars) at ($(orig.south)!0.5!(ai.north)$) {$\parallel$};
\node[right=0mm of bars] {comparison};

\end{tikzpicture}
    \caption{Illustration of the corpus generating procedure.}
    \label{fig:corpus_generation}
\end{figure}
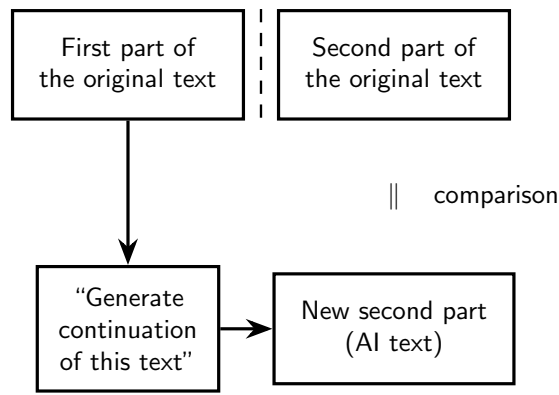

We generated the AI corpora by splitting the texts in existing corpora into two parts: the first 500 words were used as prompts, while the remaining part of the texts were used for comparison in the analyses. The basic scheme of the procedure of corpus generation is illustrated in Figure~\ref{fig:corpus_generation}. 

The main idea behind this procedure is find the position in the latent space of the model that contains given style, i.e., to give the model sufficient context to infer the style so that it can be continued.

The base models behaved as traditional language models and continued the given text chunk, without need for any additional instructions. However, the instruction tuned models had a tendency to take the text as some sort of problem to solve (answer questions from the text, summarize, evaluate or rate it etc.), so a system prompt was needed. For English, the system prompt used was: \emph{Please continue the text in the same manner and style, ensuring it contains at least five thousand words. The text does not need to be factually correct, but please make sure it fits stylistically.}

When generating Czech text, we faced language-specific issues. Some models did not respond properly to Czech system prompts, so we instead used this English system prompt: \emph{Please continue the Czech text in the same language, manner and style, ensuring it contains at least five thousand words. The text does not need to be factually correct, but please make sure it fits stylistically.}

One corpus in each language was designed to imitate the default ChatGPT persona. To do this, the system prompt reproduced the leaked ChatGPT system prompt (\cite{chatgpt_prompt}, independently confirmed by \cite{chatgpt_promptRohit} and \cite{chatgpt_promptDobos}). Because this authentic prompt does not contain explicit continuation instructions, additional instruction had to be appended after each text chunk.

\section{Analysis methods}
\subsection{Multi-dimensional analysis}
We based our analysis on the Multidimensional Analysis (MDA) to English texts using the classic framework developed by \cite{Biber1988a}, which defines six functional dimensions of register variation (see \ref{tab:biber_dimensions}) based on co-occurrence patterns among 67 lexico-grammatical features. To operationalize the original framework on our corpora, we employed Andrea Nini’s Multidimensional Analysis Tagger (MAT), a tool that replicates Biber’s tagging scheme and outputs factor scores for each text (\cite{nini2019multi}).

\begin{table}[H]
\centering
\caption{Biber’s (1988) six dimensions of register variation.}
\begin{tabular}{cll}
\hline
\textbf{Dimension} & \textbf{Positive pole} & \textbf{Negative pole} \\
\hline
1 & Involved production & Informational production \\
2 & Narrative discourse & Non-narrative discourse \\
3 & Situation-dependent reference & Explicit reference \\
4 & Overt expression of persuasion & Neutral / informational exposition \\
5 & Abstract information & Non-abstract information \\
6 & On-line elaboration & Integrated information \\
\hline
\end{tabular}

\label{tab:biber_dimensions}
\end{table}

For Czech, we adopted the multidimensional analysis model developed by \cite{Cvrceketal2020,Cvrcek.etal2018}, which is based on factor analysis over 137 linguistic features extracted from the diversified Koditex corpus (\cite{Zasinaetal2018}). This model identifies eight underlying dimensions of register variation specific to Czech (see \ref{tab:czech_dimensions}). Factor scores for each text were obtained using the MDA tagging pipeline \cite{Cvrceketal2020,Cvrcek.etal2018}, enabling the placement of Czech texts into this 8-dimensional space.

\begin{table}[H]
\centering
\caption{Dimensions of register variation in Czech (Cvrček et al., 2021).}
\begin{tabular}{cll}
\hline
\textbf{Dimension} & \textbf{Positive pole} & \textbf{Negative pole} \\
\hline
1 & dynamic & static \\
2 & spontaneous & prepared \\
3 & higher level of cohesion & lower level of cohesion \\
4 & polythematic & monothematic \\
5 & higher amount of addressee coding & lower level of addressee coding \\
6 & general / intension & particular / extension \\
7 & prospective & retrospective \\
8 & attitudinal & factual \\
\hline
\end{tabular}

\label{tab:czech_dimensions}
\end{table}

\subsection{Statistical processing}
Simple comparison in each dimension is realized as a difference between the second part of the original text chunk $\textbf{v}_{orig2}$ and the generated text chunk $\textbf{v}_{model}$ in each dimension:

\begin{equation}
\Delta \textbf{v} = \textbf{v}_{orig2}-\textbf{v}_{model}.
\end{equation}

The average of these differences for each dimension $\overline{\Delta v_d}$ can serve to compare models on each dimension separately, but cannot be used to compare the models across dimensions, since each of these dimensions have different baseline behavior (some of them differ a lot even within two halves of the same texts, others are stable). This is why the main metric of the benchmark ($b_d$) is calculated by normalizing the average differences by the standard error of differences between the two halves of the original text chunk ($i_d$), where:
\begin{equation}
\textbf{i} = \textbf{v}_{orig2}-\textbf{v}_{orig1},
\end{equation}
and
\begin{equation}
b_d = \frac{\overline{\Delta v_d}}{\textnormal{SE}(I_d)}.
\label{eq:normalized}
\end{equation}
To obtain less fine-grained, one-dimensional benchmark $B$ (only one number per model), Euclidean length of these normalized average vectors can be calculated:
\begin{equation}
B = ||\textbf{b}||.
\label{eq:length}
\end{equation}
At each step, confidence intervals are calculated by bootstrap resampling (even the SE is calculated by resampling, since the underlying distribution is unknown).

\section{Results and Discussion}
In this section, we present the main findings on stylistic variation in the AI-generated corpora. The discussion is divided into two parts: one focusing on English and the other on Czech, reflecting the fact that the MDAs were conducted separately and involved different sets of dimensions.

\subsection{Overall results}
\subsubsection{English}
\begin{figure}[p]
    \centering
    \includegraphics[width=0.88\linewidth]{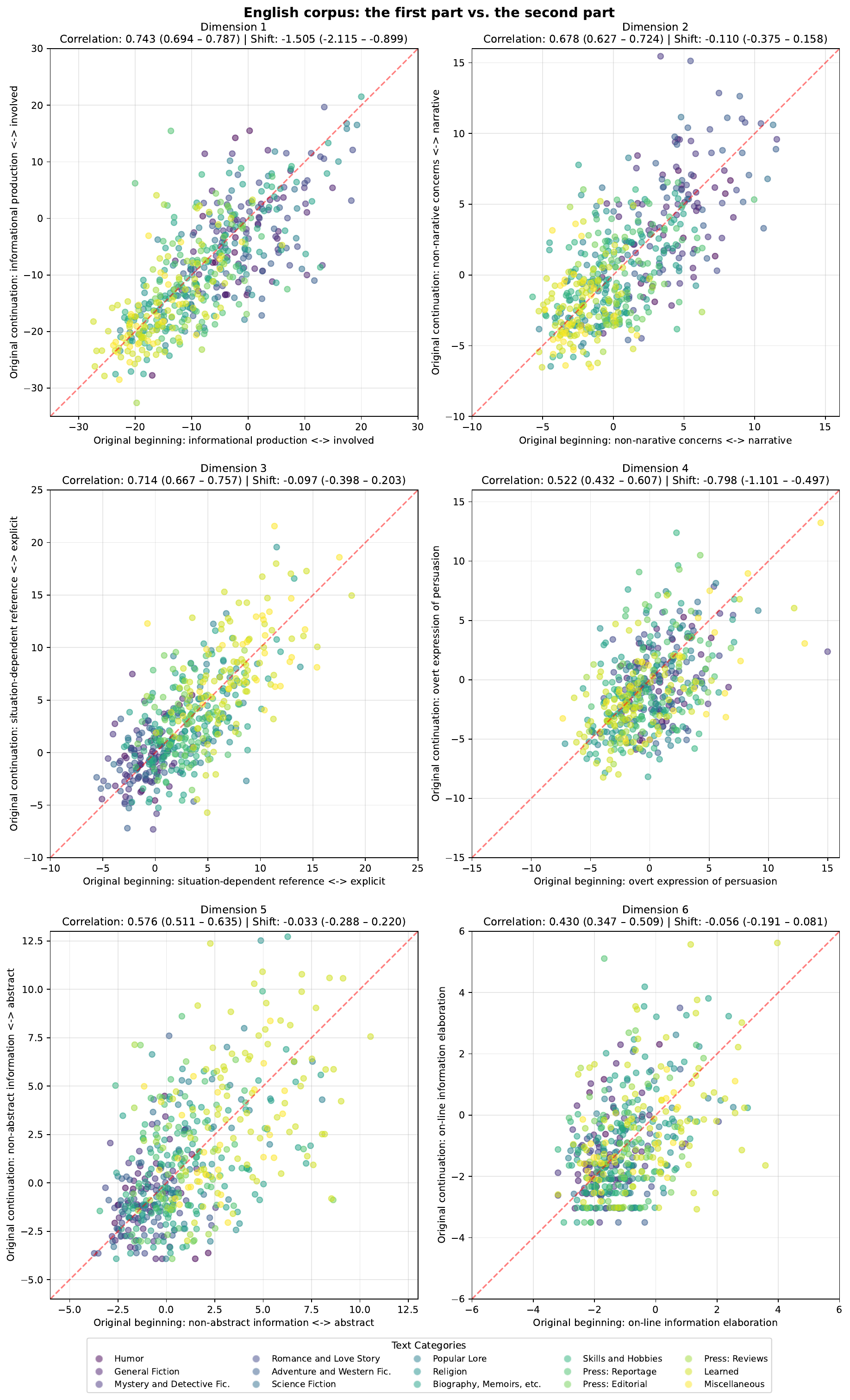}
    \caption{Relation between the first and second parts of human written text chunks across all dimensions (English).}
    \label{fig:scatter_human_English}
\end{figure}

We used scatterplots to illustrate the correlation of stylistic dimensions between the first and second parts of the human corpus (Figure~\ref{fig:scatter_human_English}). Before turning to AI-generated texts, it was important to first examine the relationship between the two parts of the original human texts. Since even within a single human-originated text the correlation is not expected to be perfect, this analysis provides a useful baseline against which to compare the AI outputs. As can be seen, the shift between the two parts of a text can itself be statistically significant (cf. confidence intervals), which suggests that we can also expect substantial differences in the case of LLMs. In other words, when interpreting model-induced shifts, we must also account for the ``stability” of human texts. For English, for example, the average shift in Dimension 1 and Dimension 4 was considerable. Dimension 1, in particular, proved relatively unstable: on average, the second parts of texts shifted by –1.5 on average, i.e. from involved production toward informational production.

\begin{figure}[p]
    \centering
    \includegraphics[width=0.88\linewidth]{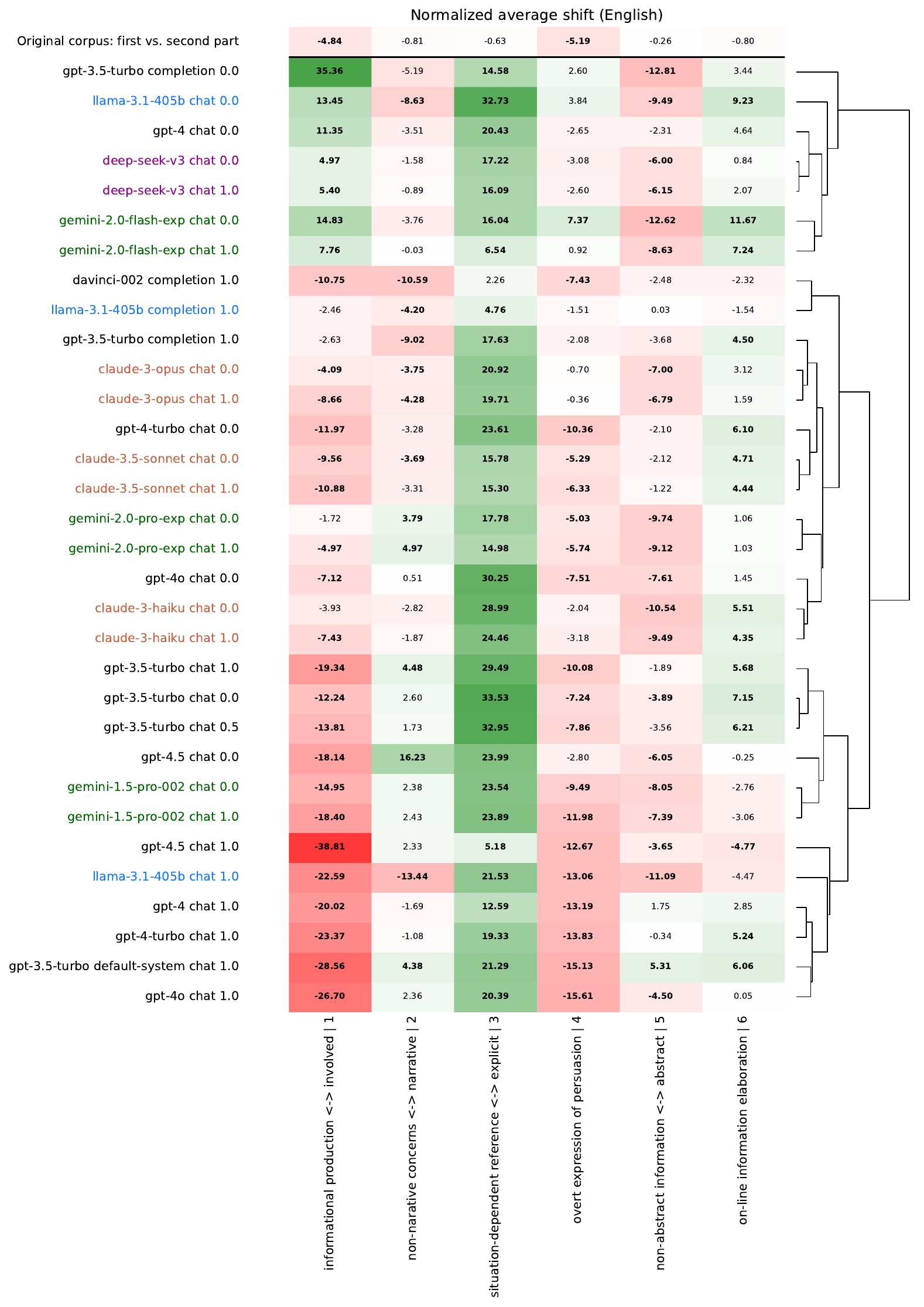}
    \caption{Normalized average shifts of each model on each dimension in English.}
    \label{fig:main_English}
\end{figure}

To see systematic patterns, Figure~\ref{fig:main_English} presents the average shifts for each model on each dimension. The dimension shift is normalized by Standard Error (SE) of the the first and second parts of the human corpus, so that the values are directly compared to the baseline (as in Formula \ref{eq:normalized}). The color coding in this chart also based of this normalization: the maximum red/green corresponds to 50 times the baseline SE. Moreover, statistically significant results are indicated in bold, adjusted with the Bonferroni correction for the actual number of tested hypotheses (in the case of English: 33 × 6).

The overview in Figure~\ref{fig:main_English} shows that although the models did not behave identically --- and some reproduced the style of the original texts more convincingly than others --- they generally tended to shift the texts in similar directions. For example, most models shifted texts toward the negative pole of Dimension 1 (from involved to informational production); on Dimension 3, from situation-dependent to explicit reference; and on Dimension 5, from abstract to non-abstract information.

The magnitude of these shifts could be quite large, but we also consistently found models that reproduced the original style very closely. This demonstrates the potential of AI language models: they are capable of imitating target styles convincingly, provided that the correct model, prompt, or decoding parameters (e.g., temperature) are chosen.

We also observed that some stylistic dimensions were generally easier for models to reproduce than others. For instance, Dimension 2 (narrative vs. non-narrative discourse) and Dimension 6 (online elaboration of information) were not shifted substantially, suggesting that these stylistic properties are relatively easy for models to imitate. By contrast, Dimension 3 showed consistent shifts toward the explicit reference pole across all models, with many exhibiting very strong shifts. This suggests that stylistic features captured by this dimension are particularly difficult for models to replicate.

The figure \ref{fig:main_English} also shows hierarchical clustering of the models according to similarity of their stylistic vectors. The dendrogram reveals that temperature seems to not play a big role: identical models just with different sampling temperature group together very frequently (e.g., deep seek v3 chat tempt 0 and 1, gemini 2.0 flash exp chat temp 0 and 1, ...). In contrast, different models from the same company usually do not cluster together, despite having similar training data or even being based on the same base model, for example GPT-4 (temperature 1) and GPT-4-turbo (temperature 1) are clustered together, as expected, but their temperature 0 counterparts are in different main clusters. All Claudes are in the same main cluster, but they are seemingly randomly mixed with other GPTs and Geminies. It would be interesting to look at model size, whether there is a tendency to cluster small models together across companies, but we do not know the exact size and we do not want to guess.

\begin{figure}[p]
    \centering
    \includegraphics[width=0.88\linewidth]{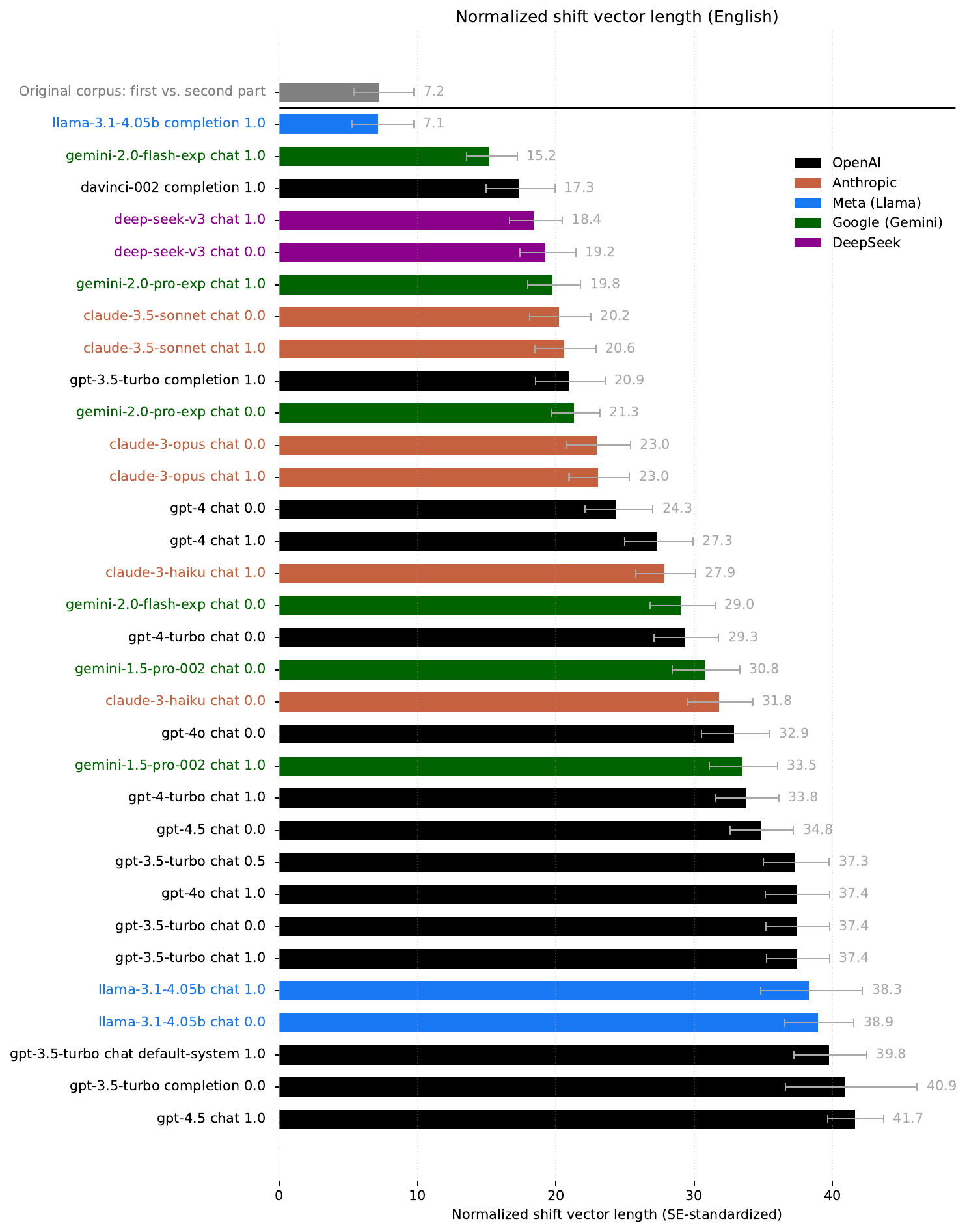}
    \caption{Normalized vector lengths for each model in English.}
    \label{fig:vector_English}
\end{figure}

The fig \ref{fig:vector_English} shows the overall score of each model, which is expressed by length of the vector (all dimensions are normalized the same way as in Figure~\ref{fig:main_English}, see Formula \ref{eq:length}). This visualization offers a more comprehensive and straightforward comparison of the models, so the readers can right away assess which model is overall stylistically the closest to the human one. However, this visualization further reduces the complexity of the stylistic variation, therefore, it should be taken just as a preliminary indication of the performance of the models.

To give readers a clearer sense of how the texts produced by ``successful” and ``less successful” models look in practice, we present an excerpt from the genre \textit{Fiction – letters, memoirs, autobiographies} (Table~\ref{tab:example_text_english}). Even in this single example, we can see that while LLaMA continued stylistically very close to the original, GPT-4.5 markedly exaggerated certain features of the style, for instance by using a large number of descriptive adjectives.

\begin{longtable}{p{0.04\textwidth}p{0.28\textwidth}p{0.28\textwidth}p{0.28\textwidth}}
\caption{Example of the original English text and its continuations by human, llama, and GPT-4.5.}
\label{tab:example_text_english}\\

\hline
\textbf{Part} & \textbf{Original} & \textbf{} & \textbf{} \\
\hline
\endfirsthead

\hline
\textbf{Part} & \textbf{Original} & \textbf{} & \textbf{} \\
\hline
\endhead

\hline
\endfoot
\endlastfoot

1/2 & \multicolumn{3}{p{0.86\textwidth}}{\textit{[...]}

\textit{This is a big change for me. I don’t think I understood hope at all before this year, even though I used the word all the time.}

\textit{“Are you going to be able to make it tonight?”
“Hope so!”
“You’ll be brilliant at this! You have nothing to worry about!”
“Hope so!”
“Everything will work out in the end, you’ll see.”
“Hope so!”}

\textit{I used it almost like an apology in advance, a managing of expectations, when I didn’t have the confidence that I could actually deliver anything I was ‘hoping’ for. I’d feel the urge}} \\[0.5em]
~&~&~&\\
& \textbf{Original} & \textbf{LLAMA} & \textbf{GPT-4.5} \\
\hline
2/2 & \vspace{0.2cm}to put it in when I couldn’t be sure something was going to happen, when I wasn’t confident something could be achieved, when there were no concrete facts available. For me, it was a ‘try’ word, a genuine longing, yes, but a wish rather than a solid promise. I realize now that that is, in fact, the opposite of how hope works. As soon as you wish for something, long for it, use hope as anything less than a certain belief, it drives the magic of hope further from you. [...] 
&
\vspace{0.2cm}to dismiss it too, like I wanted to duck out of committing to expecting anything solid or firm without knowing my chances of success. A kind of fear, really, of putting too much on the line. ‘Hope’ was such a powerful concept that I would avoid connecting with it.

The idea of hope as something tangible didn’t occur to me until I heard a doctor talking about patients’ lives being saved by a dose of hope. [...]
&
\vspace{0.2cm}to downplay commitment, to avoid disappointment. But real, deep, soul-sustaining hope — I learned — is far more than this. It’s not apologetic; instead, it is brave, grounding, life-giving.

My husband, Derek, contracted the virus early, when it felt like the whole world was still naive, perhaps even dismissive, of the scale of its threat. In the beginning, he had nothing more alarming than a persistent cough, a slight fever, and fatigue that had him sinking deeper into the sofa each day. [...]
\\

\end{longtable}



\subsubsection{Czech}
\begin{figure}[H]
    \centering
    \includegraphics[width=0.66\linewidth]{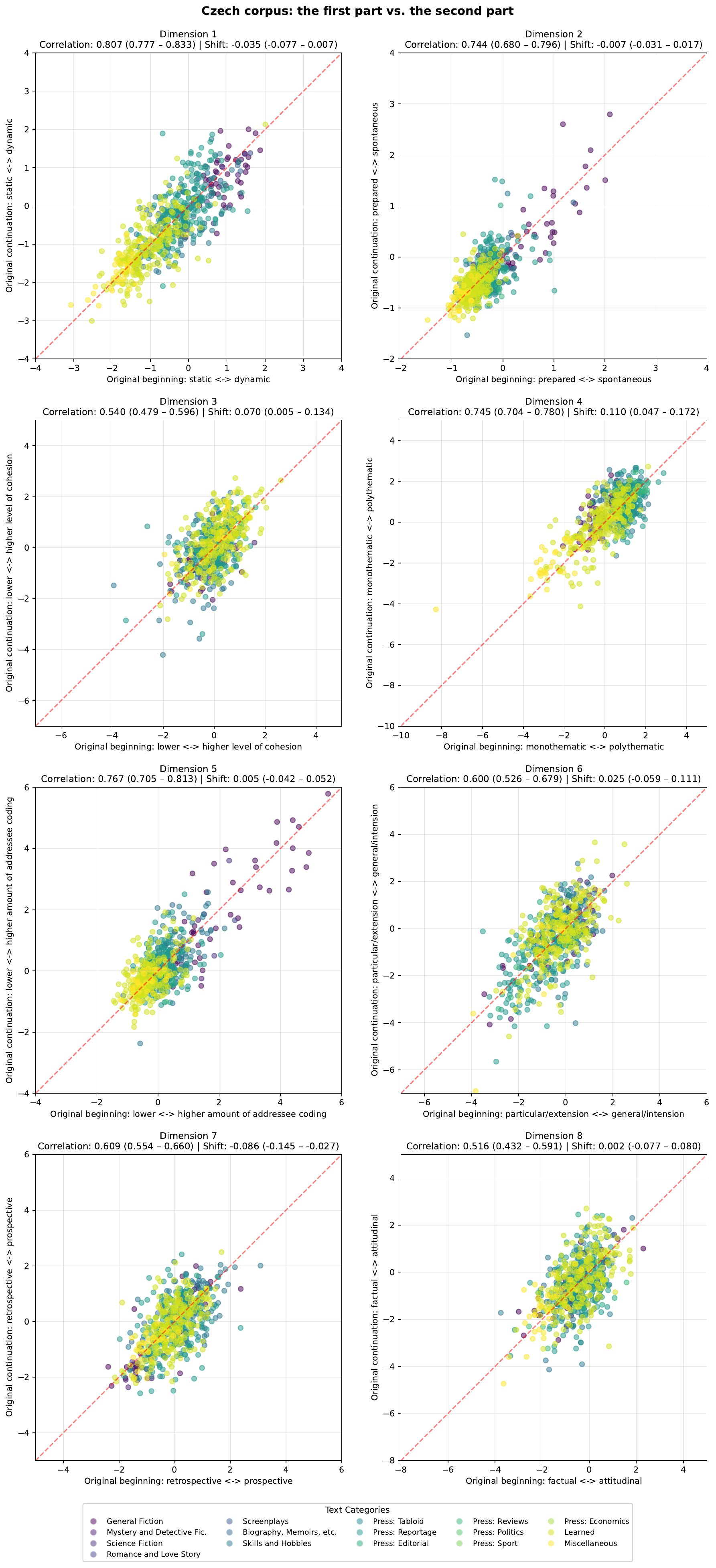}
    \caption{Relation between the first and second parts of human written text chunks across all dimensions (Czech).}
    \label{fig:scatter_human_Czech}
\end{figure}

For Czech, we proceeded in the same way as for English: first, we measured the correlation between the first and second parts of the corpus across all eight dimensions of variation (Figure~\ref{fig:scatter_human_Czech}). In comparison with English, the Czech corpus appears more ``stable,” as none of the normalized shifts between the first and second halves were statistically significant.

\begin{figure}[H]
    \centering
    \includegraphics[width=0.88\linewidth]{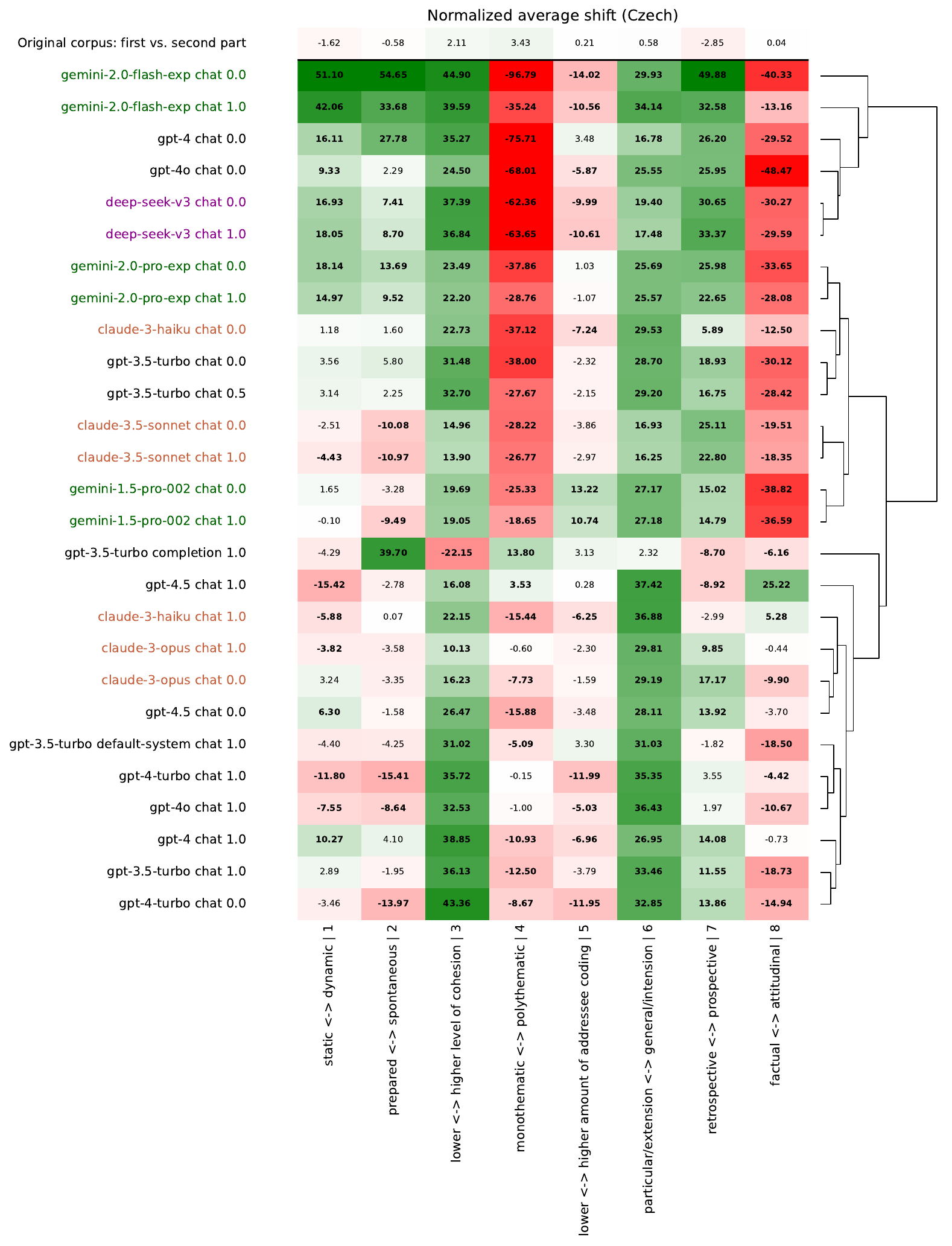}
    \caption{Normalized average shifts of each model on each dimension in Czech.}
    \label{fig:main_Czech}
\end{figure}

The Figure~\ref{fig:main_Czech} shows that compared to English, there was much higher overall shift of the texts by the models. That confirms our expectation about how models behave in English vs. in smaller languages. As expected, the models have higher problem to stylistically mimic the texts in Czech (a smaller language) than in English (a ``default" language of the majority of the training data). Some models even failed to stylistically mimic the original corpus on all eight dimensions (e.g., gemini 2.0. flash-exp chat 0.0; deep-seek-v3 chat 1.0), which did not happen in English. We also have to point out that some models are not even included in the analysis, since they were not able to generate Czech texts at all.

Similar to English, the shifts tend to move in the same direction across most models. For example, dimensions 3, 6, and 7 generally shift toward the positive poles, while dimensions 4, 5, and 8 shift toward the negative poles. The first two dimensions do not show the same consistency across models; however, the models appear to mimic these dimensions quite well, as indicated by the relatively small average shifts.

The hierarchical clustering by vector similarity (\ref{fig:main_Czech}) is, as in English, rather chaotic. Texts with the same temperature often cluster together (though not perfectly), but there is no clear system: models we would expect to be similar often are not. Some GPTs are clustered together, but not all of them and even Claudes are not in one main group.

\begin{figure}[H]
    \centering
    \includegraphics[width=0.88\linewidth]{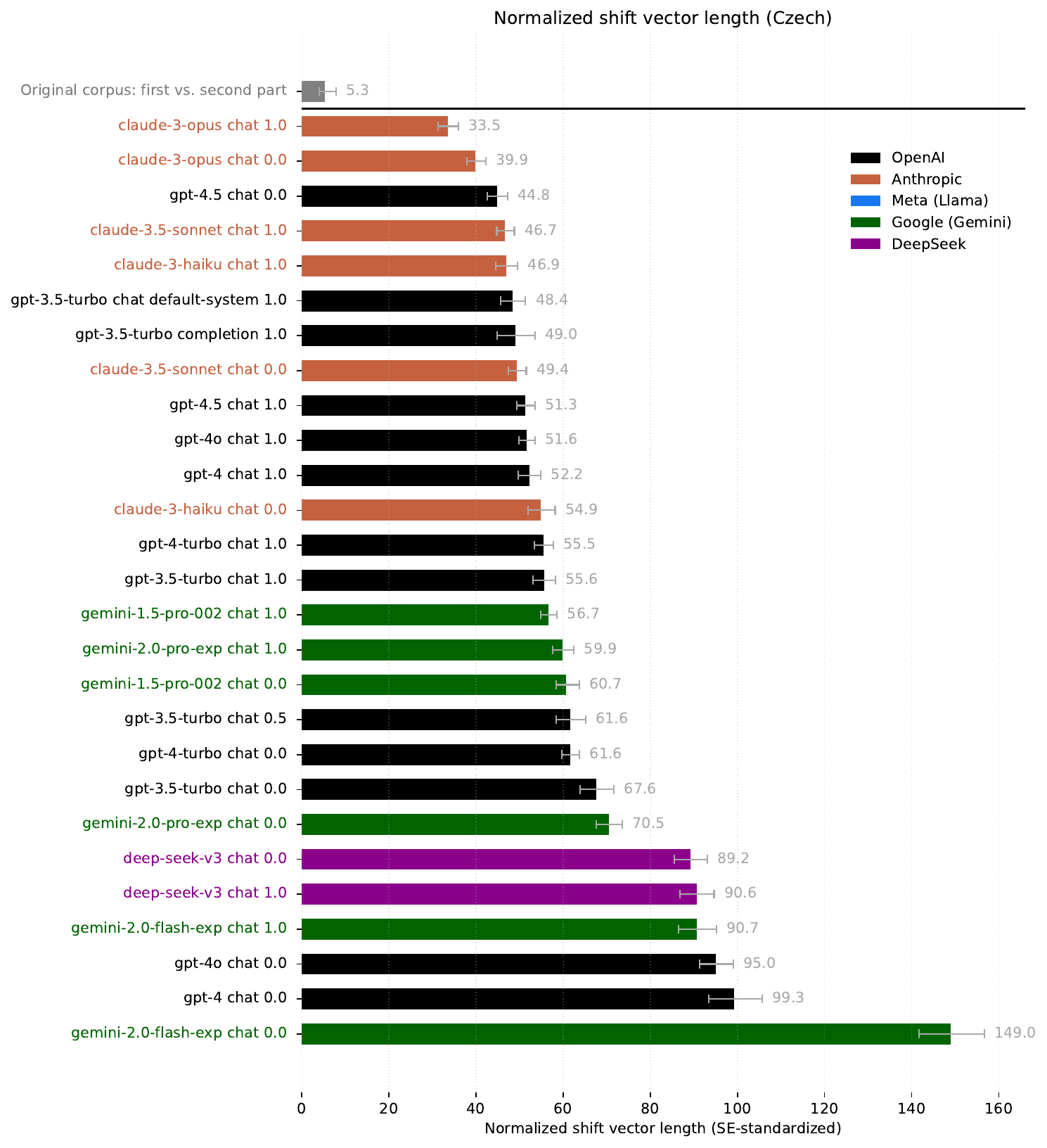}
    \caption{Normalized vector lengths for each model in Czech.}
    \label{fig:vector_Czech}
\end{figure}

Figure \ref{fig:vector_Czech} shows that, compared to English, the average vector shifts are much larger. While Llama 3.1 successfully mimicked the English corpus, achieving an average shift comparable to the difference between the first and second parts of the original corpus, no model performed as well in Czech. The best result came from Claude 3 Opus (temperature 1), which reached an average normalized shift close to 40 --- a value barely exceeded even by the worst-performing models in English.

\subsection{Base versus instruction tuned}
\subsubsection{English}
Now we will focus on one of the main questions of this article, namely, how base models differ from models that have been fine-tuned for instruction following. Interestingly, our analysis indicated that the oldest base model (davinci-002) performed very well, especially on Dimension 3, where most other models failed. This is likely because it was not fine-tuned with reinforcement learning (RLHF in the case of OpenAI, RLAIF for Anthropic) or with instruction tuning, and it was the last model not trained on synthetic data produced by earlier LLMs. However, the best overall performance was achieved by LLaMA 3.1 Completion, which is also a base model. GPT-3.5-turbo in completion mode had inconsistent results in this area; it performed very well at a sampling temperature of 1, while it had very bad results for the temperature of 0 (second to last in figure \ref{fig:vector_English}). Let us look at the details.

\begin{figure}[p]
    \centering
    \includegraphics[width=0.88\linewidth]{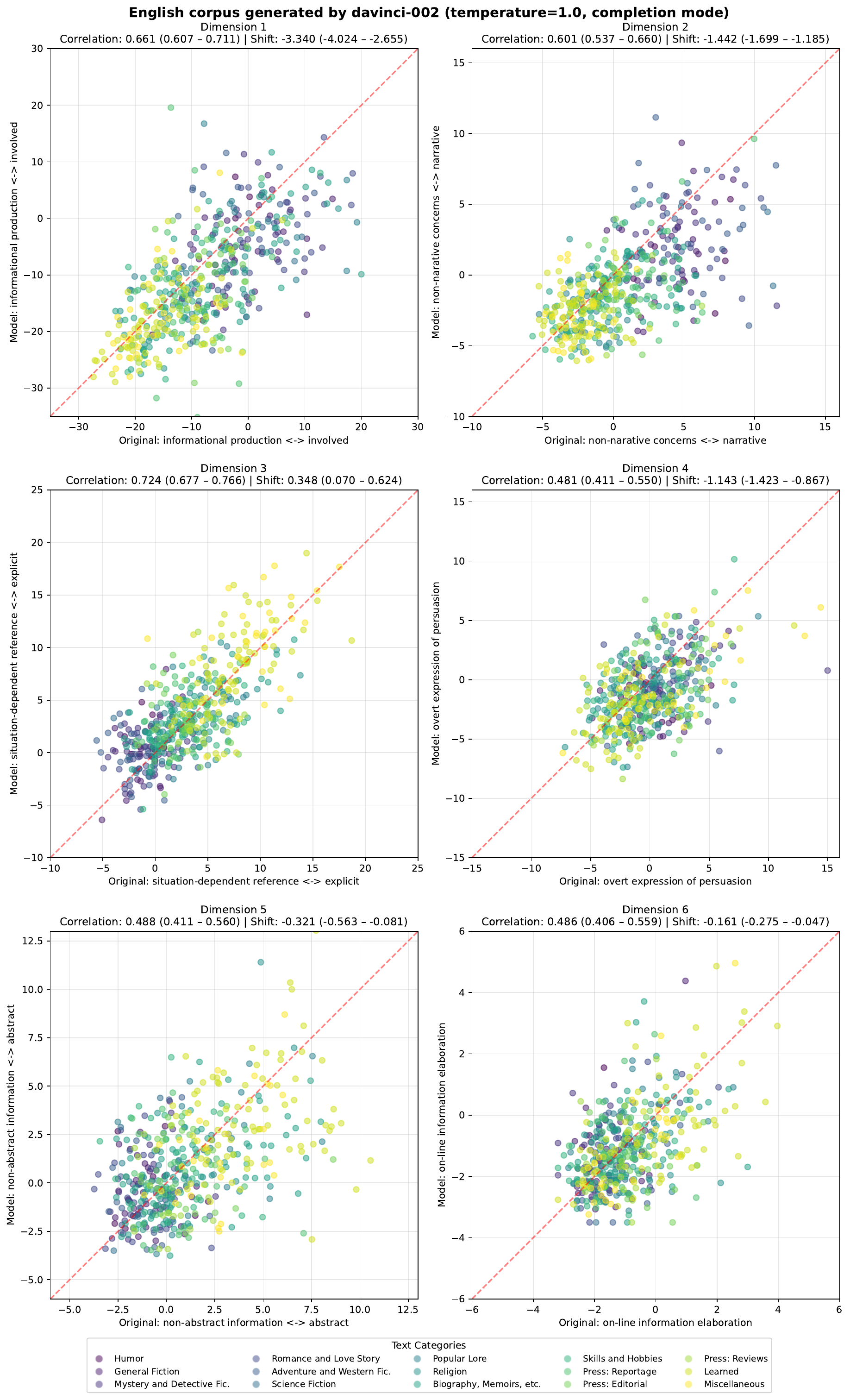}
    \caption{Relation between the second parts of human written text chunks and th second parts generated by base model Davinci-002 across all dimensions (English).}
    \label{fig:scatter_baseRLHF_English_davinci}
\end{figure}

Figure \ref{fig:scatter_baseRLHF_English_davinci} shows the shifts of texts from davinci-002 relative to the second part of the human corpus across six dimensions. The position of a text in the original corpus is plotted on the X-axis, while its position in the davinci-002 corpus is on the Y-axis. This visualization provides a more fine-grained perspective than Figure \ref{fig:main_English}, as it allows us to examine the positions of individual texts and their genre affiliations. For davinci-002, the figure shows that although all texts shifted, the dispersion remained relatively ``regular,” without noticeable outliers or strong genre effects. For the first, second, and fourth dimensions, the average shifts reached statistical significance.

\begin{figure}[p]
    \centering
    \includegraphics[width=0.88\linewidth]{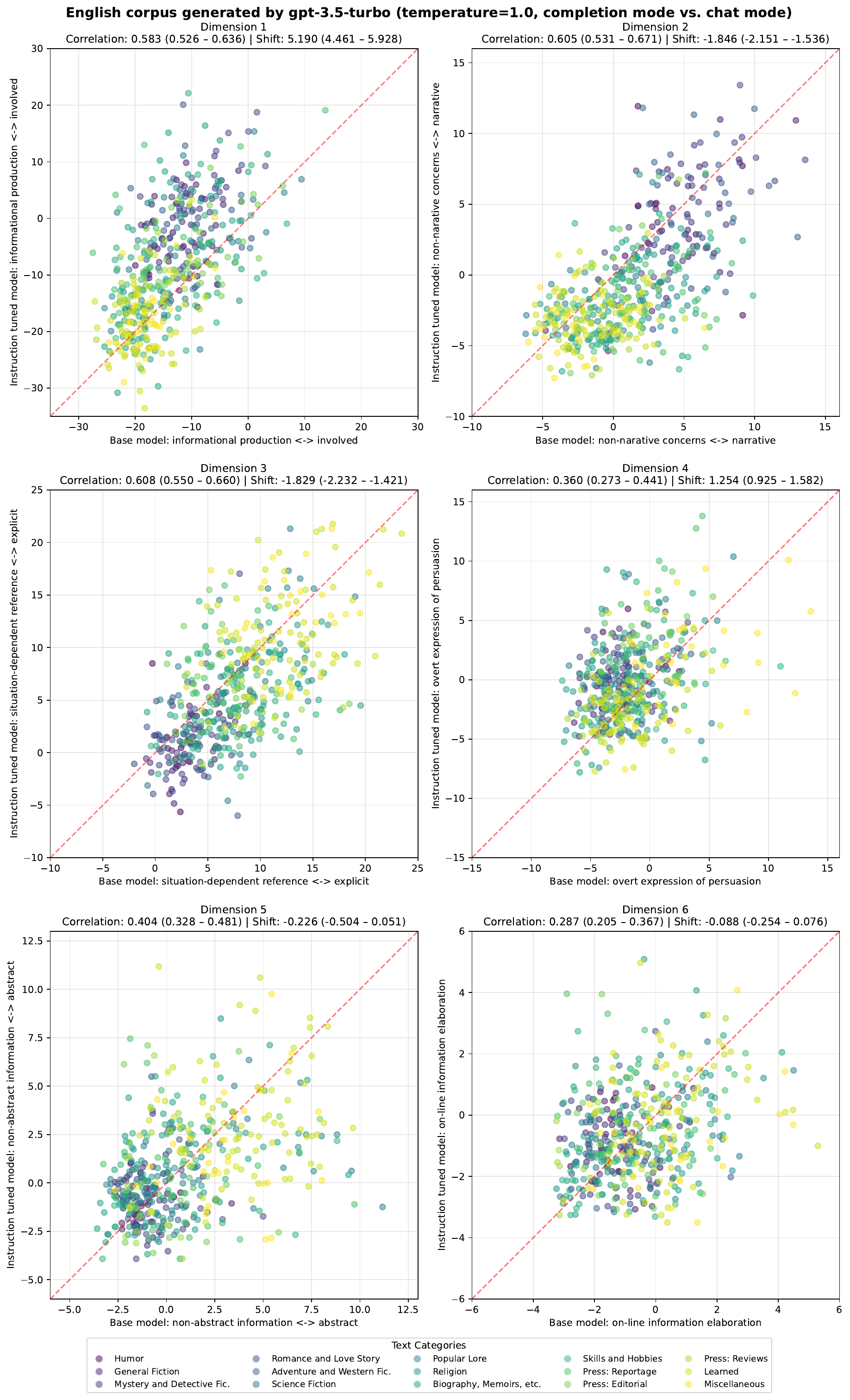}
    \caption{Relation between the second parts generated by GPT-3.5 turbo in completion mode and the second parts generated by instruction tuned GPT-3.5 turbo (English).}
    \label{fig:scatter_baseRLHF_English_GPT}
\end{figure}

In contrast, Figure \ref{fig:scatter_baseRLHF_English_GPT}, which compares the base model GPT-3.5 turbo (X-axis) with its instruction-tuned version (Y-axis), reveals a clear genre effect. The normalized shifts on the first two dimensions occurred primarily in fiction texts (humor, general, mystery, romance), while press, learned, and miscellaneous texts remained relatively stable. This is an insight we would not have gained without such a fine-grained approach. Genres associated with information production largely remained in the same region, whereas fiction --- which is typically located on the opposite pole --- became even more skewed.

\begin{figure}[p]
    \centering
    \includegraphics[width=0.88\linewidth]{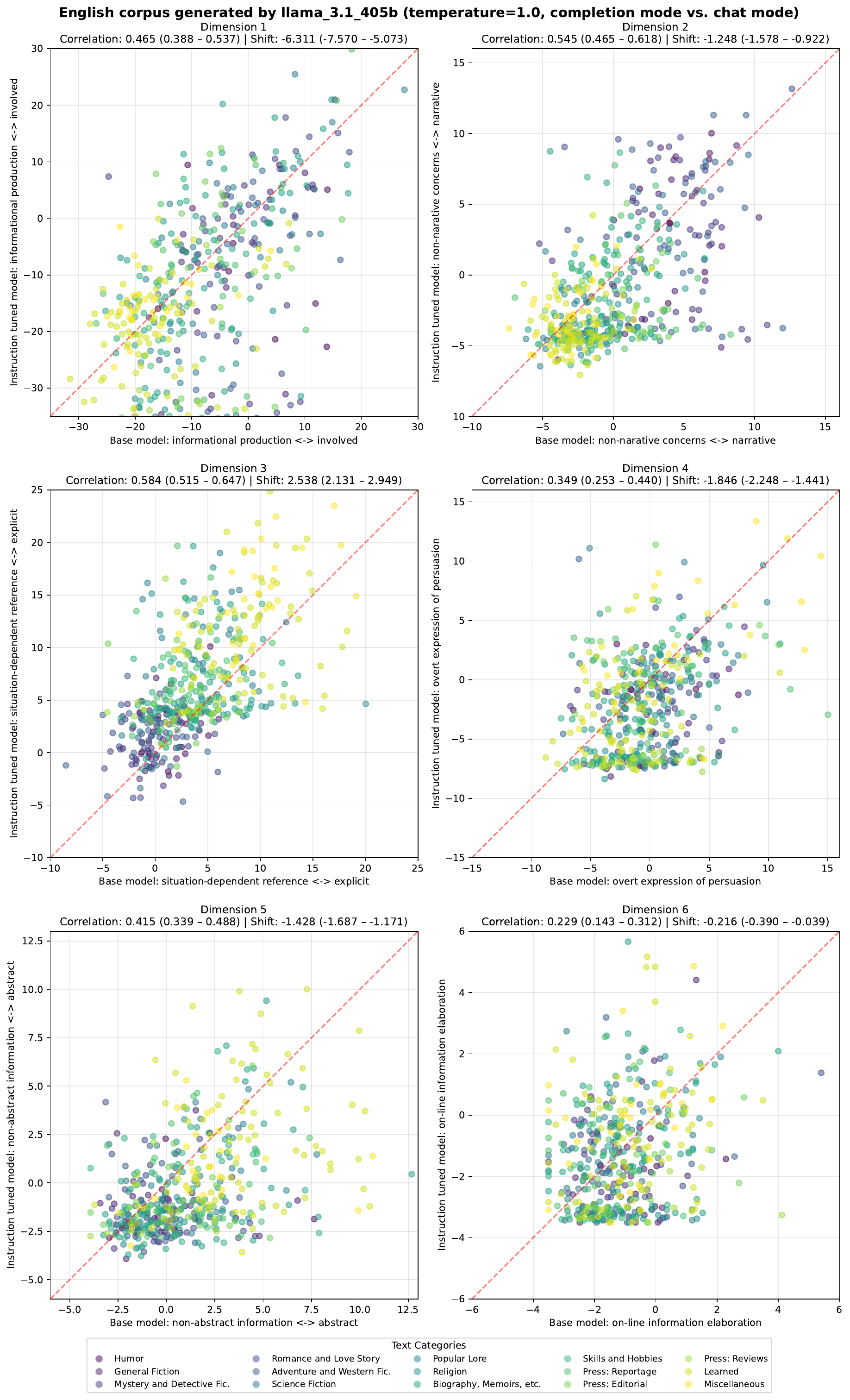}
    \caption{Relation between the second parts generated by Llama 3.1 Base and the second parts generated by instruction tuned Llama 3.1 (English).}
    \label{fig:scatter_baseRLHF_English_Llama}
\end{figure}

As already noted, Llama 3.1 generally reproduces the profile very well. However, as shown in the scatterplot in Figure~\ref{fig:scatter_baseRLHF_English_Llama}, the instruction tuned model failed completely on some texts, which are then crowded at very similar levels (around –30 on dimension 1 and around 7.5 on dimension 4).

\subsubsection{Czech}
\begin{figure}[p]
    \centering
    \includegraphics[width=0.66\linewidth]{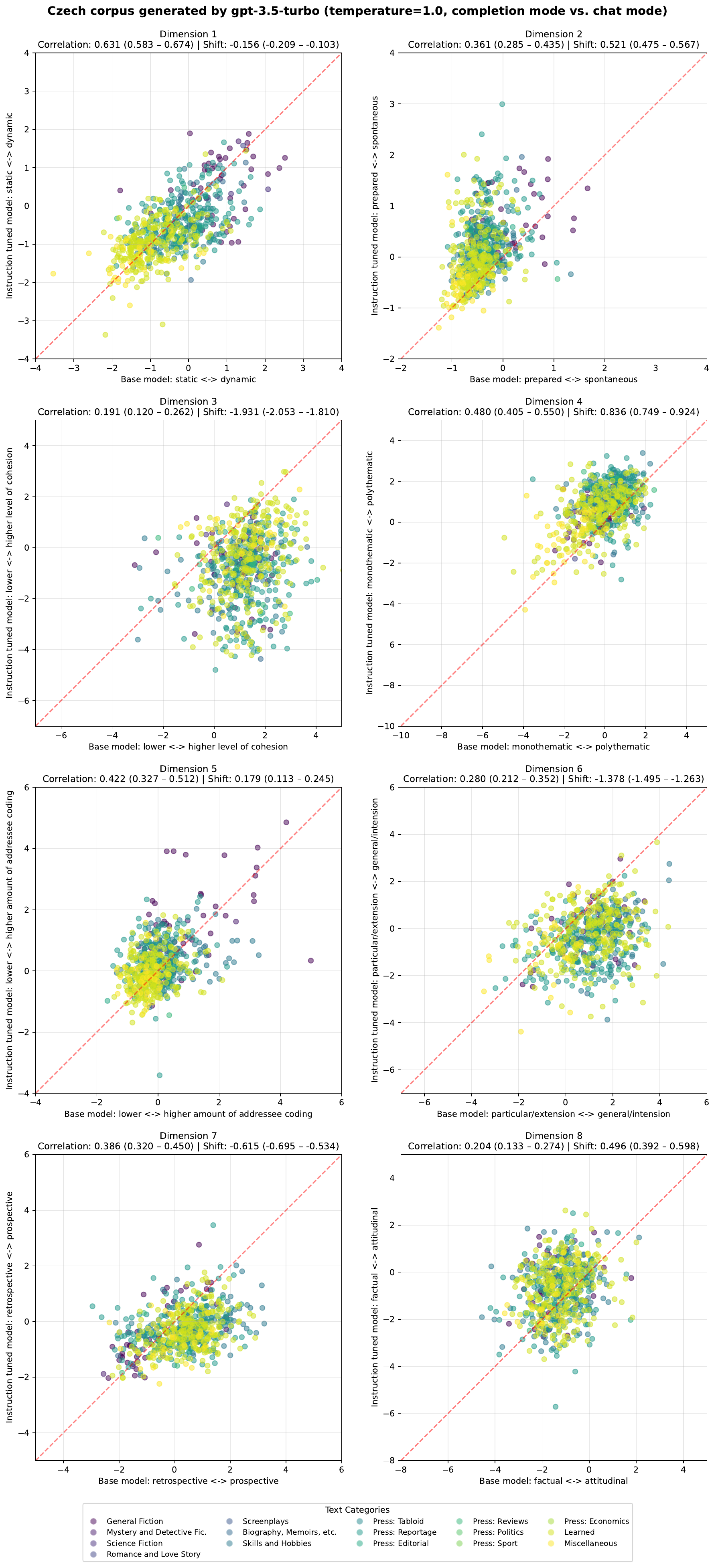}
    \caption{Relation between the second parts generated by GPT-3.5 turbo in completion mode and the second parts generated by instruction tuned GPT-3.5 turbo (Czech).}
    \label{fig:scatter_baseRLHF_Czech}
\end{figure}

In Czech, we have only one base model that can be examined in this level of detail — GPT-3.5 Turbo in completion mode — since davinci-003 did not produce reliable data in Czech, nor did Llama (the same holds true for the instruction-tuned version of Llama). This model behaved quite similarly to its chat-mode counterpart in terms of overall vector length; however, in the cluster analysis it did not show affinity with related models and thus appeared more as an outlier. This can also be seen in Figure~\ref{fig:scatter_baseRLHF_Czech}, where the effect is visible across all dimensions and genres.

Perhaps the most interesting observation for Czech is that, in fact, no truly functional base model exists for this language.

\subsection{Prompt impact} 
\subsubsection{English}
\begin{figure}[p]
    \centering
    \includegraphics[width=0.88\linewidth]{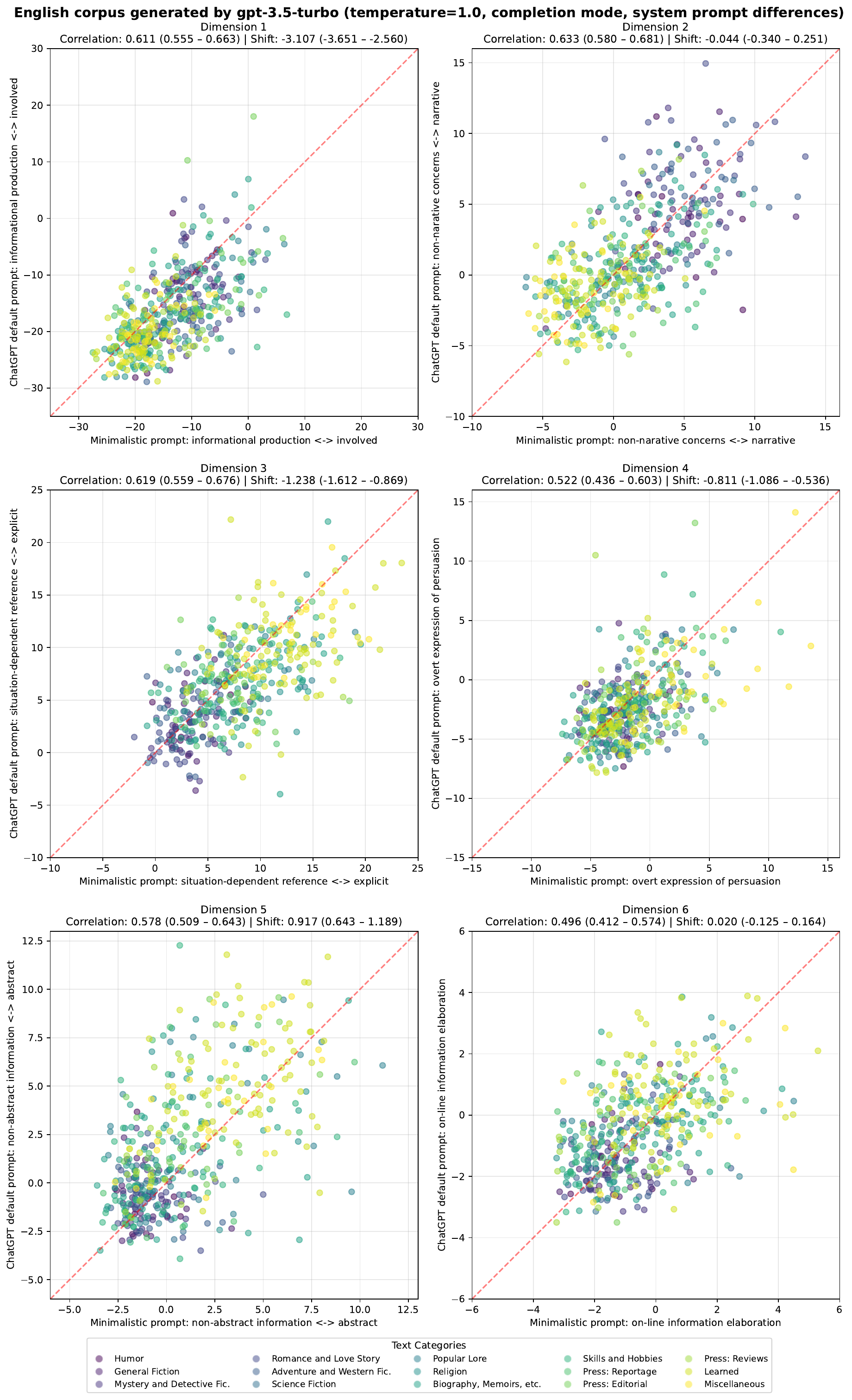}
    \caption{Relation between the second parts generated by GPT-3.5 turbo with short system prompt and the second parts generated by GPT-3.5 turbo with default ChatGPT system prompt (English).}
    \label{fig:scatter_prompt_English}
\end{figure}

In the same way as we examined the difference between base and instruction-tuned models, we now look at the differences between texts produced by a minimalist system prompt and texts that were produced using the original ChatGPT long system prompt. These texts were produced using the GPT-3.5-turbo model, which is the model with which this prompt was used in November 2023. In Figure~\ref{fig:scatter_prompt_English}, the x-axis shows values for texts with minimalist prompt ``Please continue..."), while the y-axis shows those with the long prompt (``You are ChatGPT..."), and as can be seen, the shifts are relatively uniform across text types, although the overall shift is quite large. These large shifts are also noticeable in the cluster analysis (Figure~\ref{fig:main_English}), where GPT-3.5 turbo with the long prompt clusters more with newer models GPT-4 turbo, GPT-4, and GPT-4o than with the original GPT-3.5 turbo. We can therefore conclude that in English, the difference in personas is quite large.

\subsubsection{Czech}
\begin{figure}[p]
    \centering
    \includegraphics[width=0.66\linewidth]{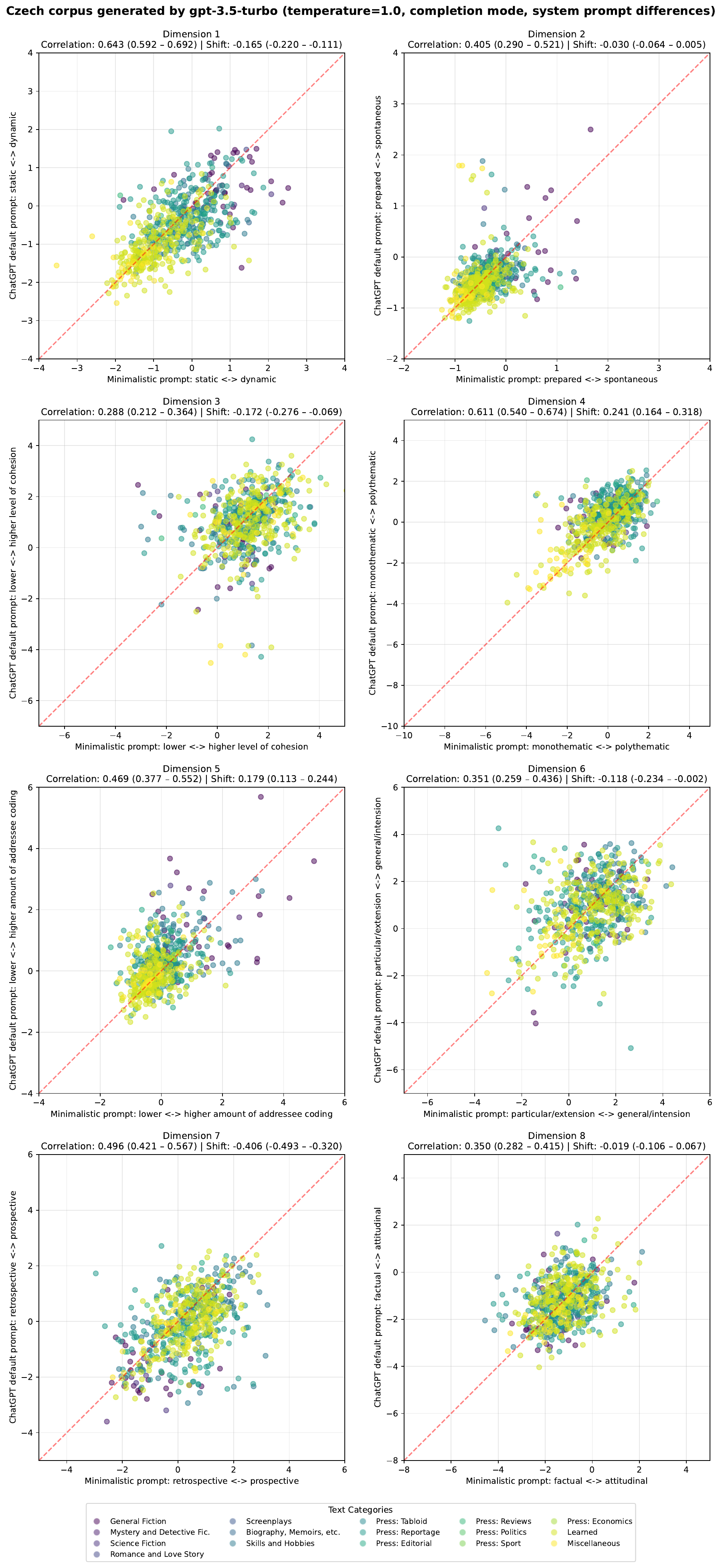}
    \caption{Relation between the second parts generated by GPT-3.5 turbo with short system prompt and the second parts generated by GPT-3.5 turbo with default ChatGPT system prompt (Czech).}
    \label{fig:scatter_prompt_Czech}
\end{figure}

A similarly uniform shift can be seen for the same model in Czech (Figure~\ref{fig:scatter_prompt_Czech}). Although the long prompt is in English, so we might expect some more pronounced and more skewed results, for Czech results the performance compared to other models is relatively good (in the first quarter of the results ranking, while in English it is at the end, see Figure~\ref{fig:vector_Czech}). Interestingly, even in Czech it clusters with the same models as in English, namely GPT-4 turbo and GPT-4o, though in this case it was at least placed alongside GPT-3.5 turbo with the short prompt within the larger cluster (Figure~\ref{fig:main_Czech}).

\subsection{Temperature impact}
We did not find a systematic effect of temperature on the results in either language. As described earlier, Figures~\ref{fig:main_Czech} and~\ref{fig:main_English} show that the same model, when sampled with different temperature settings, still clustered together according to vector length. This suggests that temperature does not have a major influence on stylistic variation. We should note, however, that temperature can have a noticeable effect on semantics. When the temperature is set to 0, the model behaves deterministically, always choosing the most probable token, which results in highly consistent but less diverse outputs. By contrast, a temperature of 1 introduces greater randomness in token selection, leading to more varied and creative generations.

In principle, such variation could influence MDA features related to lexical diversity (e.g., moving average type–token ratio). However, even if differences between temperatures 0 and 1 occurred in this respect, they did not translate into systematic differences at the level of multidimensional analysis.


\section{Conclusion}
In this article, we explored stylistic variation in LLMs with the aim of establishing a replicable stylistic benchmark. We developed a reproducible method for measuring the \emph{stylistic shift} of LLM-generated language relative to human-produced language.  

Our comparative method builds on the classical Biberian approach to linguistic variation, employing established stylistic dimensions that have been corroborated in the linguistic literature over several decades. We measured the position of text chunks produced by LLMs on these dimensions and compared them to the positions of comparable chunks written by humans. The average differences were then normalized using the standard error of differences between two consecutive human-produced text chunks, which allowed us to account for the natural variance present in human texts.  

This normalization against two human text samples is essential. Human texts themselves are heterogeneous, and if an AI continuation proved markedly more homogeneous than its human counterparts, this would also signal unsuccessful simulation of human language. 

The benchmark operates on three complementary levels of granularity:

\begin{enumerate}
    \item \textbf{Position of individual texts on the stylistic dimensions.}  

    This level is useful for analyzing what happens within each model on the scale of single texts. Since we have metadata on text types, we can observe genre-specific tendencies and identify outliers, which may then be examined qualitatively. However, such fine-grained analysis makes direct model-to-model comparison difficult, as the volume of information requires statistical testing of differences at the corpus level.

    \item \textbf{Corpus-level averages per dimension.}  

    At this level, the benchmark is expressed as a single value for each stylistic dimension for the entire model, accompanied by confidence intervals. This makes it possible to compare multiple models directly, while still preserving a multi-layered perspective that allows for meaningful interpretation. For instance, we may find that one model exhibits substantially higher narrativity than another.

    \item \textbf{Vector length.}  

    The most coarse-grained metric, representing overall performance: how far a model’s stylistic profile diverges from human text. This is calculated as the Euclidean length of the normalized stylistic shift vector.  
\end{enumerate}

We applied this benchmark to heterogeneous corpora constructed from a wide range of large language models. This design enabled us to address the following research questions, the answers to which we summarize below.

    \emph{How well can current LLMs produce stylistically diverse texts from various genres and text types?}

LLMs can achieve this task fairly successfully, but there are substantial differences between the models. With respect to genres, our analysis of dimensional shifts shows that certain text types are easier for models to imitate than others. Thus, if the goal of a user is to obtain a text of a particular genre that seems as written by a human, it is often insufficient to rely on the most popular or widely used model at a given time. Instead, one must sometimes explore different models, try different prompts, etc., something that most users do not do. This lack of exploration may contribute to common misconceptions about what AI can and cannot achieve (see, e.g., \cite{milička2025humanslearndetectaigenerated}).

    \emph{Are texts created using current LLMs stylistically shifted consistently across different models? I.e., is there some AI-language stylistic attractor?}

We found evidence of certain stylistic attractors toward which most models converge. In English, for instance, some dimensions showed relatively little average shift across models—such as Dimension 2 (narrative vs. non-narrative discourse) and Dimension 6 (online elaboration of information) — indicating that these stylistic properties are comparatively easy for models to reproduce. By contrast, Dimension 3 exhibited consistent and often strong shifts toward the explicit reference pole, suggesting that features captured by this dimension are particularly difficult for LLMs to replicate.  

In Czech, we observed similar patterns: most models showed shifts in the same direction across dimensions. Specifically, Dimensions 3, 6, and 7 tended to shift toward the positive pole, while Dimensions 4, 5, and 8 shifted toward the negative pole. The first two dimensions showed no consistent directional bias, yet models generally imitated them well, as indicated by the small average shifts.  

Overall, there do appear to be general ``AI tendencies.” In our framework these are expressed as dimensions, while in other studies they are described as particular linguistic features.   

    \emph{How do instruction-tuned models differ from base models?}

Base models often performed stylistically better than instruction-tuned ones, though this varies by language. In English, base models such as davinci-002 or LLaMA 3.1 produce stylistically human-like texts. This observation is in accordance with findings by (\cite{reinhart2025llms}). In Czech, however, most base models could not be used at all, as they failed to generate coherent text (either repeating fragments or reverting to English).  

    \emph{What is the difference between texts generated using a simple system prompt and texts generated using long ``helpful assistant" system prompt?}
For both English and Czech, the differences are significant in both statistic and real-world sense. The stylistic profile of GPT-3.5 turbo with the original Chat-GPT system prompt is less similar to the same model with the minimal system prompt than to newer Open-AI models.
    
    \emph{Are stylistic features dependent on the sampling temperature?}   

We did not find systematic effects of sampling temperature. Outputs generated with temperature 0 and with temperature 1 typically clustered together and showed similar stylistic profiles.
    
    \emph{Is the stylistic shift smaller in English than in a language underrepresented in the training data?}
    
English showed better outcomes than Czech in several respects: it showed smaller stylistic shifts compare to human texts overall; there are more models that successfully imitate human language across more dimensions; and there is a greater model availability in general. This finding points to a broader conclusion: the extreme default orientation of LLMs toward English (in pre-training data, alignment processes, etc.) reduces the competitiveness of smaller languages.

We mean: increasingly more texts are co-written with assistance of AI, we can expect that this tendency will increase. We can assume that the languages where the AI tools perform the best will benefit most. English, already dominant, might get even stronger support, while smaller languages risk being sidelined. There’s a danger of a feedback loop: weaker LLM support leads to less use, which leads to fewer examples and less training data, which further weakens support. In the long run, that could create a stylistic divide, where English texts come closer to human-like variation, while underrepresented languages drift further into more ``AI-specific" styles.

For future research, our study should be replicated with other underrepresented languages to test the generalization of the findings for Czech. In addition, it will be important to continue monitoring the performance of new models in this benchmark. As our results show, newer or larger models do not necessarily score better in terms of stylistic imitation of human language compare to older and smaller models.

\section{Data availability}
The figures in vector graphics format and fine-grained results for all models that could not be presented here, as well as the scripts that were used to analyze the data, are available at \url{https://osf.io/hs7xt}. The original corpora (Koditex and BE21) cannot be made available for copyright reasons, but  they can be searched on \url{https://www.korpus.cz/kontext/query?corpname=koditex_v2}  (Koditex), and \url{https://cqpweb.lancs.ac.uk/} (BE21). The AI-generated corpora can be fully downloaded from \url{http://milicka.cz/kezstazeni/aicorpora.zip}, and can be also searched on KonText (\url{https://korpus.cz/kontext}) as AI Brown and AI Koditex.

More results and interactive charts are available on the official website of the benchmark (\url{https://korpus.cz/stylisticbenchmark/}).

\section{Acknowledgements}
We thank the participants at the ICAME 2024 (Vigo) conference and the ISCA/ITG workshop (Berlin) for stimulating discussion when we presented the research.

\section{Funding}
Jiří Milička was supported by Czech Science Foundation Grant No. 24-11725S, gacr.cz (Large language models through the prism of corpus linguistics).

This work was supported by the project “Human-centred AI for a Sustainable and Adaptive Society” (reg. no.: CZ.02.01.01/00/23\_025/0008691), co-funded by the European Union.

Anna Marklová was supported by Primus Grant PRIMUS/25/SSH/010.

\section{Declaration on using AI}
The GPT-4, GPT-4o, GPT-4.5, and GPT-5 models by OpenAI, and Claude 4 Sonnet and Claude 4.1 Opus models by Anthropic were consulted for coding scripts and language editing of the article. However, all scripts and texts underwent manual review and were, when necessary, corrected or further refined, and the authors assume full responsibility for any errors.

\section*{References}
\printbibliography[heading=none]

\end{document}